\RequirePackage{tikz}
\documentclass[sn-mathphys,Numbered]{sn-jnl}%
\usepackage{graphicx}%
\usepackage{multirow}%
\usepackage{amsmath,amssymb,amsfonts}%
\usepackage{amsthm}%
\usepackage{mathrsfs}%
\usepackage[title]{appendix}%
\usepackage{xcolor}%
\usepackage{textcomp}%
\usepackage{manyfoot}%
\usepackage{booktabs}%
\usepackage{algorithm}%
\usepackage{algorithmicx}%
\usepackage{algpseudocode}%
\usepackage{listings}%

\DeclareMathOperator*{\argmax}{\arg\max}

\begin{document}

\title[A Novel Memetic Strategy for Optimized Learning of Classification Trees]{A Novel Memetic Strategy for Optimized Learning of Classification Trees}

\author{\fnm{Tommaso} \sur{Aldinucci} \email{tommaso.aldinucci@unifi.it}}

\affil{\orgdiv{Dipartimento di Ingegneria dell'Informazione}, \orgname{University of Florence}, \orgaddress{\street{Via di Santa Marta 3}, \city{Florence}, \postcode{50139}, \country{Italy}}}


\abstract{Given the increasing interest in interpretable machine learning, classification trees have again attracted the
attention of the scientific community because of their glass-box structure. These models are usually built
using greedy procedures, solving subproblems to find cuts in the feature space that minimize some impurity
measures. In contrast to this standard greedy approach and to the recent advances in the definition of the
learning problem through MILP-based exact formulations, in this paper we propose a novel evolutionary
algorithm for the induction of classification trees that exploits a memetic approach that is able to handle datasets with thousands of points. Our procedure combines the exploration of the feasible space of solutions
with local searches to obtain structures with generalization capabilities that are competitive with the state-of-the-art methods.}


\maketitle

\section{Introduction}\label{sec1}
In the context of supervised learning, Classification Trees (CTs) are some of the most widely used models for classification problems \cite{kotsiantis2013decision,song2015decision} especially for tabular data.
A CT is a connected and acyclic graph where each internal node (branch) performs a split in the feature space to divide data using some learned conditions. Finally, terminal nodes (leaves) are responsible for providing class predictions.

Most of the known algorithms for inducing decision trees perform univariate splits (axis-aligned) \cite{breiman84classification, quinlan1986induction} i.e., they choose a single feature which is used to compare the value of the corresponding component of each data point with respect to a threshold. In this way, the structure obtained can be easily interpreted by a domain expert since each prediction of the model is completely determined through the path followed by the point in the tree i.e., through decision rules like ``if then else'' type.
Thus, it's not surprising that this kind of models are very common and useful especially in the healthcare context \cite{intrator1992decision, olanow2001algorithm, bertsimas2019development} and every time the model understanding is a major requirement \cite{Rudin2019Stop}. 
To this aim, shallow trees, having a manageable number of rules, are a good trade-off between model expressiveness and complexity of explanation, indeed, it's well known that if allowed to grow large, these classifiers tend to lose their interpretability and to overfit.

Focusing on the interpretability importance, in this paper we address the problem of the induction of shallow axis-aligned classification trees by proposing \textit{Tree Memetic Optimization} (TMO), the first (to the best of the authors' knowledge) genetic algorithm for axis-aligned classification trees induction which employs a memetic \cite{moscato1989evolution} methodology which acts at a tree level. 
Our method exploits a standard evolutionary technique to explore the space of feasible classification trees of a given depth and the memetic approach to maintain a population of classifier with good predictive performance.
We empirically show that, for shallow trees, our evolutionary algorithm, taking advantage of local searches, is able to combine exploration and exploitation to search for good models in the combinatorial space of feasible solutions. For this reason it can be a valid alternative to standard greedy approaches, post refinement strategies or exact formulations of the learning problem.

To assess the quality of our method we used three different baselines: the most common greedy algorithm CART \cite{breiman84classification}, the Tree Alternating Optimization (TAO) method proposed in \cite{carreira2018alternating} and the recent exact model, OCT, based on Mixed Integer Linear Programming \cite{bertsimas2017optimal}.
Unlike MILP formulations that do not scale well and often fail to find better solutions than the one provided in warm starts, our approach can handle datasets of thousands of points and it is able to obtain models with competitive performance.

The rest of the manuscript is organized as follows:
\begin{itemize}
    \item In section \ref{sec2} we discuss the main works for classification trees learning that have similarities with our proposal;
    \item In section \ref{sec3} we give a detailed description of our method;
    \item In section \ref{sec4} we report and discuss numerical comparison between our method and the baselines;
    \item In section \ref{sec5} we summarize our contribution and we suggest some possible future developments.
\end{itemize}

\section{Related Works}\label{sec2}
The induction of optimal decision trees is a $\mathcal{NP}$-complete problem \cite{laurent1976constructing}. Because of this,  most algorithms are based on top-down greedy strategies using quality measures such as Gini impurity \cite{breiman84classification}, information gain \cite{quinlan1986induction} and distance measures \cite{de1991distance}.
Although easily applicable, heuristic methods usually lead to suboptimal structures. In addition, the recursive nature of these strategies induces partitions that make difficult the attribute selection at deeper nodes, generating overfitting. For this reason, top-down greedy methods often require a post pruning phase which employs complexity measures to 
achieve trees with better generalization properties, but actually making the whole training process in two phases.

Different methods have been proposed to overcome these issues. First of all, the sub-optimal structure induced by the greedy growth led to the need for new algorithms.
Over the years, the interest in formulating the decision tree learning problem through linear programming has been increasing. Some examples of this, although with different purposes, can already be found in the early '90s when linear methods for constructing separation hyperplanes and category discrimination has been proposed \cite{bennett1992robust, bennett1994multicategory}.

Also linear programming methods for the induction of Decision Trees have been explored \cite{bennett1992decision, bennett1994global}.
However, despite the formulations, these types of approaches have never been considered for practical purposes.
In recent years, due to the improvement in the hardware capabilities and also in the calculation performance of solvers, new formulations based on Mixed Integer Linear Programming (MILP) of the problem have been developed both for univariate and multivariate splits \cite{bertsimas2017optimal, verwer2019learning, bessiere2009minimising, gunluk2021optimal}.
Some works are also focused on the interpretability of CTs, introducing regularization terms both for the number of branch nodes \cite{bertsimas2017optimal} and for the total number of leaves in the model \cite{hu2019optimal, lin2020generalized}.
Although these formulations are exact, they often introduces a huge number of binary variables, making the model not suitable with real world datasets.

Other approaches that attempt to induce the global structure of the tree, without guarantee of optimality, involve the use of Evolutionary Algorithms (EAs) and Genetic Algorithms (GAs) \cite{eiben2003introduction}.

The general idea of these methods is based on principles of natural selection and genetics \cite{holland1992adaptation}. Starting from a population (set) of individuals (solutions), at each iteration a new solution is defined through selection and cross-over operations. The quality of the obtained candidate is then assessed by means of a fitness function, thus deciding whether it can replace another solution in the current set \cite{jones1998genetic}. 
Due to the absence of assumptions about the objective function, these strategies have proven to be very effective in solving global optimization problems and over the years many algorithms of this type have been developed \cite{storn1997differential, kennedy1995particle, dorigo2006ant}.
Given the popularity of these techniques, evolutionary approaches have also been proposed both for data mining problems and for knowledge discovery \cite{freitas2003survey}.

Also in the context of decision tree learning, genetic algorithms have been employed. Instead of building the model by exploiting a greedy strategy i.e. by solving subproblems iteratively, this family of algorithms performs a more robust search in the global space of solutions \cite{barros2011survey}.
In particular, in \cite{cantu2003inducing} a top-down method for inducing oblique decision trees is presented which uses an evolutionary approach to solve the hyperplane selection problem at each node, showing that their strategy can build oblique trees whose accuracy is competitive with respect to other  methods. 
In \cite{papagelis2001breeding} is proposed a population-based evolutionary method that dynamically evolves a set of minimal decision trees using a fitness function which is balanced between the accuracy on the known instances and the complexity of the new solution.

An extension of evolutionary algorithms is the memetic variant \cite{moscato1989evolution}.
This technique acts by performing local searches or refinement methods to improve the quality of new solutions.
Memetic strategies are quite common in the field of global optimization \cite{jia2011effective, wu2019memetic} and, recently, they have been also proposed for solving clustering problems \cite{mansueto2021memetic}.

Given the empirical effectiveness of this approach, the memetic variant has also been investigated for decision trees in \cite{kretowski2008memetic, czajkowski2012does}. However, local searches employed in these early papers operate at an inner nodes level, greedily exploiting node information during the mutation step. The complete tree structure is never considered as a whole during the optimization process.
We deem that the local optimization phase has to be run once a new solution has been defined and,
for this reason, we propose to use local searches on the whole tree, using a specialized optimizer as TAO, encouraging the exploitation of local optimization for the evolution of better solutions.

\section{Proposed method}\label{sec3}
In this section we describe our memetic algorithm, TMO, in the context of binary classification problems.
For any $n \in \mathbb{N}$, let $D : = \{ (x_i, y_i), \ x_i \in \mathbb{R}^p , \ y_i \in \{0, 1 \} , \ i = 1,..., n \}$ be a finite set of data points with $p$ features. To alleviate the combinatorial explosion of the feasible space, we search for good classification trees of a given maximum depth $d$, avoiding to introduce further complexity in the exploration of the space of feasible tree structures.

The first phase of any evolutionary algorithm is the initialization of the population, i.e, the definition of an initial set of solutions. The most common method to do this is through random sampling in the feasible space, however, we deem that a totally random initialization may not be a good choice in the case of a population of decision trees because these models have a strong structural dependency by construction. Thus, a random sampling initialization may cause the algorithm to explore "dead zones" of the search space, slowing the convergence to near-optimal solutions. 

Since it is clear that there is a trade-off between exploration and exploitation \cite{vcrepinvsek2013exploration}, to provide enough diversity without compromising the quality of the individuals, our approach employs a Random Forest \cite{breiman2001} to initialize the population. This choice is motivated by bagging \cite{breiman1996bagging} and the random subspace method \cite{ho1998random} which are used during the training of the Random Forest to ensure a sufficient heterogeneity in the ensemble.

Thus, starting with a Random Forest $RF := \{ t_1, ..., t_k \}$ of $k$ decision trees as population, our algorithm performs an initial optimization phase on each model using the \textit{Tree Alternating Optimization} (TAO) algorithm \cite{carreira2018alternating} which exploits an alternating minimization technique on all the levels of the decision tree. Doing this, every model of the initial population becomes a local optima in the feasible space of classification trees for the given maximum depth $d$.

After this first step, a standard genetic algorithm requires to carry out the evolution implementing two main phases: selection and crossover.
Selection is the process by which individuals from the population are chosen to participate in the definition of a new solution. Usually, this is carried out by favouring individuals with high fitness, in the belief that they may contain good components for generating better solutions \cite{goldberg1991comparative}.
The selection technique greatly affects the convergence rate and, in general, an higher selection pressure results in higher convergence rates but it increases the chances that the population may converge prematurely to a local optimum.
For this reason, we decided to use a selection method which is quite similar to the one in \cite{storn1997differential} that is, we do not take into account the fitness of the individuals. 
Thus, our method iterates on each model and it samples another one from the population to obtain the couple of parents that will define a new solution.
Let $i \in \{1,..., k\} $ the index of the current tree, a second index $j$ is sampled from a uniform distribution $\mathcal{U}_{\{j=1,..., k, \ j \neq i \}}$ over all the possible indexes except $i$.
The solutions $t_i$ and $t_j$ are then encoded through a one-to-one mapping in a fixed-length list. These new representations will be used by the stochastic crossover operation, detailed in the next section, to define a new candidate $\hat{t}$.

\begin{algorithm}[h]
    \caption{TMO scheme} 
    \label{alg:tmo}
    \begin{algorithmic}
        \State \textbf{Input}: Train data $D$, initial population $ P = \{{t_1,...,t_k}\}$
        \State $f_{best} = \max \{f(t), \ t \in P \}$
        \State $t_{best} = \argmax \{f(t), \ t \in P \}$
        \For {$g = 1,.., n_{gen}$}
            \State Get a bootstrap sample $B_D^{(g)}$
            \For{$i = 1 ,..., k$}
                \State Sample $j \sim \mathcal{U}_{\{j=1,..., k, \ j \neq i \}}$    
                \For{node $z = 1,..., 2^d - 1$}
                    \If{$\mathcal{U}_{[0, 1]} < CR $}
                        \State $\hat{t}^z = t_j^z$
                    \Else
                        \State $\hat{t}^z = t_i^z$
                    \EndIf
                \EndFor
                \State Optimize $\hat{t}$ using $\mathcal{L}(\hat{t} \ ; B_D^{(g)})$
                \If {$f(\hat{t}) > f(t_i)$}
                    \State $t_{i} = \hat{t}$
                \EndIf
                \If {$f(\hat{t}) > f_{best}$}
                    \State $f_{best} = f(\hat{t})$
                    \State $t_{best} = \hat{t}$
                \EndIf
            \EndFor
        \EndFor
    \end{algorithmic}
\end{algorithm}

Unlike common genetic strategies, before asserting the quality of the new solution through a fitness function, our method exploits the memetic strategy to locally optimize the new candidate with respect to the misclassification loss using TAO.
This local optimization phase, indicated as $\mathcal{L}(\hat{t} \ ; B_D)$, is made with respect to a bootstrap sample $B_D$ of the training data. This allows to maintain a sufficient diversity, avoiding an over-intensive exploitation that would compromise the algorithm's ability to explore the feasible space and to generate structures with good generalization performance.
Finally, the quality of the improved candidate $\hat{t}$ is assesses by means of the accuracy on the train set which acts as a fitness function.
The pseudo-code of our method is reported in Algorithm \ref{alg:tmo}.

\subsection{Tree encoding scheme}

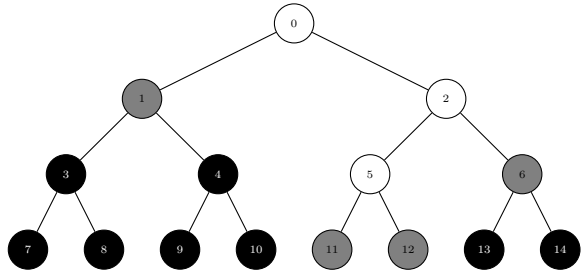
\begin{figure}[h]
    
    \begin{center}
        \begin{tikzpicture}[ level distance=1.0cm,
          level 1/.style={sibling distance=4cm},
          level 2/.style={sibling distance=2cm},
          level 3/.style={sibling distance=1.0cm}]
        \tikzstyle{every node}=[font=\footnotesize]
        \node[circle, draw, minimum width=0.518cm](0){\scalebox{.6}{0}}
          child{
            node[circle,draw, minimum width=0.518cm, fill=gray, text = black ](1){\scalebox{.6}{1}}
            child{
                node[circle,draw, minimum width=0.518cm, fill=black, text=white ](3) {\scalebox{.6}{3}}
                child{node[circle,draw, minimum width=0.518cm, fill=black, text = white] {\scalebox{.6}{7}}}
                child{node[circle,draw, minimum width=0.518cm, fill=black, text = white] {\scalebox{.6}{8}}}} 
            child{
                node[circle,draw, minimum width=0.518cm, fill=black, text=white](4) {\scalebox{.6}{4}}
                child{node[circle,draw, minimum width=0.518cm, fill=black, text = white] {\scalebox{.6}{9}}}
                child{node[circle,draw, minimum width=0.518cm, fill=black, text = white] {\scalebox{.6}{10}}}} 
            } 
          child{
            node[circle,draw, minimum width=0.518cm]{\scalebox{.6}{2}} 
            child{
                node[circle,draw, minimum width=0.518cm, fill = white, text = black](5) {\scalebox{.6}{5}}
                child{node[circle,draw, minimum width=0.518cm, fill=gray, text = black] {\scalebox{.6}{11}}}
                child{node[circle,draw, minimum width=0.518cm, fill=gray, text = black] {\scalebox{.6}{12}}} 
            }
            child{
                node[circle,draw, minimum width=0.518cm, fill=gray](6) {\scalebox{.6}{6}}
                child{node[circle,draw, minimum width=0.518cm, fill=black, text=white] {\scalebox{.6}{13}}}
                child{node[circle,draw, minimum width=0.518cm, fill=black, text=white] {\scalebox{.6}{14}}}
            }
            };
        \end{tikzpicture}
    \end{center}
    
    \vspace{1em}
    \caption{An unbalanced tree of depth 3}
    \label{fig:Tree1}
\end{figure}
One of the key problems in the context of GAs applied to decision trees is the encoding mechanism used to get useful representations of feasible solutions. Different strategies can be found in the literature to address this problem but, in general, most common approaches used a tree-based encoding \cite{barros2011survey}.

Our method maps each tree to a fixed-length list of tuples containing the parameters of every node in the structure. Since the class associated with each leaf is always chosen by majority from the labels of the points arriving on the leaf itself, these nodes do not directly contribute to the definitions of new solutions, thus the last level of the structure can be omitted from the representation without loss of information.

Similarly to the method in \cite{aitkenhead2008co}, given a tree $t$ of depth $d$, the encoding mechanism considers a balanced structure of depth $d - 1$ and it uses a tuple of two elements to represent each node. 
Since the information of a branch $z$ is completely defined by the feature index $f_z$ and the threshold value $\tau_z$ that the node uses to perform the split, we use the tuple $(f_z, \tau_z)$ to encode the parameters of this kind of nodes and the conventions $(-1, -1)$, $(nil, nil)$ for leaves and missing nodes respectively.

The $2^d-1$-length encoding list is finally built using all the representation tuples and it is sorted following the order that would be used by a Breadth-First Search (BFS) to visit the tree, giving priority to the left child.

Figure \ref{fig:Tree1} shows an example of an unbalanced tree with respect to a complete structure of depth 3. Branches, leaves and missing nodes are on white, gray and black backgrounds respectively. The encoding of this structure, using our method, is the 7-length list defined as:

$$[(f_0, \tau_0), (-1, -1), (f_2, \tau_2), (nil, nil), (nil, nil), (f_5, \tau_5), (-1, -1)]$$

\subsection{Generating new solutions}
The core of every evolutionary algorithm is the method by which the definition of new solutions occurs. This phase usually is defined through the crossover operation. Our method employs a stochastic variant of this procedure, encouraging the exploration of the feasible space. After the selection step, the parents are encoded using the strategy described in the previous paragraph and the crossover operation is performed on each node till depth $d-1$. More precisely, the nodes of the new solution are defined with respect to a cross-rate parameter $CR$. 
Given $i, j$ as the indexes of the parents, each node $z$ of the new solution $\hat{t}$ is defined as:

\vspace{1em}
\begin{equation*}
\hat{t}^z =
    \begin{cases}
        t_j^z & \text{if} \ \ \mathcal{U}_{[0, 1]} \leq CR\\
        t_i^z & \text{otherwise}
    \end{cases}
\end{equation*}
\vspace{1em}

Thus, we set each node of the new solution to have the same parameters as the second parent node with probability $CR$ and to have those of the first parent with probability $1 - CR$.
This operation is obviously trivial to implement using the fixed-length encoding of each model. However, it is able to produce radical modifications in the structure of the new solution with respect to the parents.
\begin{figure}[ht]
    
    \begin{center}
        \begin{tikzpicture}[ level distance=1.0cm,
          level 1/.style={sibling distance=4cm},
          level 2/.style={sibling distance=2cm},
          level 3/.style={sibling distance=1.0cm}]
        \tikzstyle{every node}=[font=\footnotesize]
        \node[circle, draw, minimum width=0.518cm](0){\scalebox{.6}{0}}
          child{
            node[circle,draw, minimum width=0.518cm, fill=gray, text = black ](1){\scalebox{.6}{1}}
            child{
                node[circle,draw, minimum width=0.518cm, fill=black, text=white ](3) {\scalebox{.6}{3}}
                child{node[circle,draw, minimum width=0.518cm, fill=black, text = white] {\scalebox{.6}{7}}}
                child{node[circle,draw, minimum width=0.518cm, fill=black, text = white] {\scalebox{.6}{8}}}} 
            child{
                node[circle,draw, minimum width=0.518cm, fill=black, text=white](4) {\scalebox{.6}{4}}
                child{node[circle,draw, minimum width=0.518cm, fill=black, text = white] {\scalebox{.6}{9}}}
                child{node[circle,draw, minimum width=0.518cm, fill=black, text = white] {\scalebox{.6}{10}}}} 
            } 
          child{
            node[circle,draw, minimum width=0.518cm]{\scalebox{.6}{2}} 
            child{
                node[circle,draw, minimum width=0.518cm, fill = gray, text = black](5) {\scalebox{.6}{5}}
                child{node[circle,draw, minimum width=0.518cm, fill=black, text = white] {\scalebox{.6}{11}}}
                child{node[circle,draw, minimum width=0.518cm, fill=black, text = white] {\scalebox{.6}{12}}} 
            }
            child{
                node[circle,draw, minimum width=0.518cm, fill=gray](6) {\scalebox{.6}{6}}
                child{node[circle,draw, minimum width=0.518cm, fill=black, text=white] {\scalebox{.6}{13}}}
                child{node[circle,draw, minimum width=0.518cm, fill=black, text=white] {\scalebox{.6}{14}}}
            }
            };
        \end{tikzpicture}
    \end{center}
    
    \vspace{1em}
    \caption{Modification \textit{Branch/Leaf} on node 5 with respect to the structure in figure \ref{fig:Tree1}}
    \label{fig:Tree2}
\end{figure}

\begin{figure}[ht]
    
    \begin{center}
        \begin{tikzpicture}[ level distance=1.0cm,
          level 1/.style={sibling distance=4cm},
          level 2/.style={sibling distance=2cm},
          level 3/.style={sibling distance=1.0cm}]
        \tikzstyle{every node}=[font=\footnotesize]
        \node[circle, draw, minimum width=0.518cm](0){\scalebox{.6}{0}}
          child{
            node[circle,draw, minimum width=0.518cm, fill=white, text = black ](1){\scalebox{.6}{1}}
            child{
                node[circle,draw, minimum width=0.518cm, fill=gray, text=black ](3) {\scalebox{.6}{3}}
                child{node[circle,draw, minimum width=0.518cm, fill=black, text = white] {\scalebox{.6}{7}}}
                child{node[circle,draw, minimum width=0.518cm, fill=black, text = white] {\scalebox{.6}{8}}}} 
            child{
                node[circle,draw, minimum width=0.518cm, fill=gray, text=black](4) {\scalebox{.6}{4}}
                child{node[circle,draw, minimum width=0.518cm, fill=black, text = white] {\scalebox{.6}{9}}}
                child{node[circle,draw, minimum width=0.518cm, fill=black, text = white] {\scalebox{.6}{10}}}} 
            } 
          child{
            node[circle,draw, minimum width=0.518cm]{\scalebox{.6}{2}} 
            child{
                node[circle,draw, minimum width=0.518cm, fill = white, text = black](5) {\scalebox{.6}{5}}
                child{node[circle,draw, minimum width=0.518cm, fill=gray, text = black] {\scalebox{.6}{11}}}
                child{node[circle,draw, minimum width=0.518cm, fill=gray, text = black] {\scalebox{.6}{12}}} 
            }
            child{
                node[circle,draw, minimum width=0.518cm, fill=gray](6) {\scalebox{.6}{6}}
                child{node[circle,draw, minimum width=0.518cm, fill=black, text=white] {\scalebox{.6}{13}}}
                child{node[circle,draw, minimum width=0.518cm, fill=black, text=white] {\scalebox{.6}{14}}}
            }
            };
        \end{tikzpicture}
    \end{center}
    
    \vspace{1em}
    \caption{Modification \textit{Leaf/Branch} on node 1 with respect to the structure in figure \ref{fig:Tree1}}
    \label{fig:Tree3}
\end{figure}
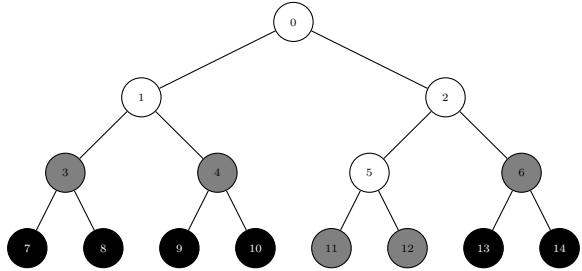

\begin{figure}[ht]

    \begin{center}
        \begin{tikzpicture}[ level distance=1.0cm,
          level 1/.style={sibling distance=4cm},
          level 2/.style={sibling distance=2cm},
          level 3/.style={sibling distance=1.0cm}]
        \tikzstyle{every node}=[font=\footnotesize]
        \node[circle, draw, minimum width=0.518cm](0){\scalebox{.6}{0}}
          child{
            node[circle,draw, minimum width=0.518cm, fill=gray, text = black ](1){\scalebox{.6}{1}}
            child{
                node[circle,draw, minimum width=0.518cm, fill=black, text=white ](3) {\scalebox{.6}{3}}
                child{node[circle,draw, minimum width=0.518cm, fill=black, text = white] {\scalebox{.6}{7}}}
                child{node[circle,draw, minimum width=0.518cm, fill=black, text = white] {\scalebox{.6}{8}}}} 
            child{
                node[circle,draw, minimum width=0.518cm, fill=black, text=white](4) {\scalebox{.6}{4}}
                child{node[circle,draw, minimum width=0.518cm, fill=black, text = white] {\scalebox{.6}{9}}}
                child{node[circle,draw, minimum width=0.518cm, fill=black, text = white] {\scalebox{.6}{10}}}} 
            } 
          child{
            node[circle,draw, minimum width=0.518cm, fill = gray, text = black]{\scalebox{.6}{2}} 
            child{
                node[circle,draw, minimum width=0.518cm, fill = black, text = white](5) {\scalebox{.6}{5}}
                child{node[circle,draw, minimum width=0.518cm, fill=black, text = white] {\scalebox{.6}{11}}}
                child{node[circle,draw, minimum width=0.518cm, fill=black, text = white] {\scalebox{.6}{12}}} 
            }
            child{
                node[circle,draw, minimum width=0.518cm, fill=black, text = white](6) {\scalebox{.6}{6}}
                child{node[circle,draw, minimum width=0.518cm, fill=black, text=white] {\scalebox{.6}{13}}}
                child{node[circle,draw, minimum width=0.518cm, fill=black, text=white] {\scalebox{.6}{14}}}
            }
            };
        \end{tikzpicture}
    \end{center}
    
    \vspace{1em}
    \caption{Modification \textit{Branch/NIL} on node 5 with respect to the structure in figure \ref{fig:Tree1}}
    \label{fig:Tree4}
\end{figure}

\begin{figure}[ht]
    \begin{center}
        \begin{tikzpicture}[ level distance=1.0cm,
          level 1/.style={sibling distance=4cm},
          level 2/.style={sibling distance=2cm},
          level 3/.style={sibling distance=1.0cm}]
        \tikzstyle{every node}=[font=\footnotesize]
        \node[circle, draw, minimum width=0.518cm](0){\scalebox{.6}{0}}
          child{
            node[circle,draw, minimum width=0.518cm, fill=white, text = black ](1){\scalebox{.6}{1}}
            child{
                node[circle,draw, minimum width=0.518cm, fill=white, text=black ](3) {\scalebox{.6}{3}}
                child{node[circle,draw, minimum width=0.518cm, fill=gray, text = black] {\scalebox{.6}{7}}}
                child{node[circle,draw, minimum width=0.518cm, fill=gray, text = black] {\scalebox{.6}{8}}}} 
            child{
                node[circle,draw, minimum width=0.518cm, fill=gray, text=black](4) {\scalebox{.6}{4}}
                child{node[circle,draw, minimum width=0.518cm, fill=black, text = white] {\scalebox{.6}{9}}}
                child{node[circle,draw, minimum width=0.518cm, fill=black, text = white] {\scalebox{.6}{10}}}} 
            } 
          child{
            node[circle,draw, minimum width=0.518cm]{\scalebox{.6}{2}} 
            child{
                node[circle,draw, minimum width=0.518cm, fill = white, text = black](5) {\scalebox{.6}{5}}
                child{node[circle,draw, minimum width=0.518cm, fill=gray, text = black] {\scalebox{.6}{11}}}
                child{node[circle,draw, minimum width=0.518cm, fill=gray, text = black] {\scalebox{.6}{12}}} 
            }
            child{
                node[circle,draw, minimum width=0.518cm, fill=gray](6) {\scalebox{.6}{6}}
                child{node[circle,draw, minimum width=0.518cm, fill=black, text=white] {\scalebox{.6}{13}}}
                child{node[circle,draw, minimum width=0.518cm, fill=black, text=white] {\scalebox{.6}{14}}}
            }
            };
        \end{tikzpicture}
    \end{center}
    
    \vspace{1em}
    \caption{Modification \textit{NIL/Branch} on node 3 with respect to the structure in figure \ref{fig:Tree1}}
    \label{fig:Tree5}
\end{figure}

Figures \ref{fig:Tree2}, \ref{fig:Tree3}, \ref{fig:Tree4}, \ref{fig:Tree5} show some example of this. Let $t_i$ be the reference tree in figure \ref{fig:Tree1}, a branch to leaf modification on node 5 is reported in figure \ref{fig:Tree2}. The effect of this operation is to prune all the sub-tree having the node 5 as root, setting this node as leaf with class obtained by majority with respect to the labels of the points arriving at that node. 

On the other hand, the inverse operation is reported in figure \ref{fig:Tree3}. In this case the leaf node 1 becomes a branch with the same parameters of the other parent and its two leaves are created.

Because in the population there may exist very unbalanced structures, modifications like nil/branch and branch/nil are also possible. In those cases the structure of the new solution undergoes the most massive transformations. 

In figure \ref{fig:Tree4} a branch/nil alteration on node 5 is shown. Because the node has to be delated from the structure, this implies that its father has to become a leaf. Thus, this kind of transformation acts as a total pruning of the sub-trees rooted at that node and at the brother (node 6), setting their father as leaf. Note that the same result can also be achieved with a branch/leaf change applied to the parent node. 

Figure \ref{fig:Tree5} shows a nil/branch conversion on node 3. This operation enlarges the structure of the model. Since a new branch has to be created, also the parent node (node 1), being a leaf, has to become a branch to maintain a correct hierarchy in the model. The parameters of the parent node are initialized as random while those of the current node are taken from the other tree participating in the crossover. Finally, the brother node (node 4) has also to be created and it is set to be a leaf with the class chosen by majority respect to the labels of the points arriving to the leaf itself.

\section{Numerical experiments}\label{sec4}

\begin{table}[ht]
    \centering
    \resizebox{\textwidth}{!}{
    \begin{tabular}{lccccc}
    \textbf{Dataset} & \textbf{Dimension} & \textbf{CART} & \textbf{TAO} & \textbf{TMO} & \textbf{OCT} \\ [0.5pt]
    \hline \\ [1pt]
    A2a        & 2265 $\times$ 119 & 81.02 $\pm$ 1.57 & 81.02 $\pm$ 1.57 & 80.93 $\pm$ 1.25 & 78.81 $\pm$ 3.56 \\ [3pt]
    A4a        & 4781 $\times$ 122 & 82.28 $\pm$ 0.77 & 82.28 $\pm$ 0.77 & \textbf{82.34} $\pm$ \textbf{0.74} & 82.28 $\pm$ 0.77 \\ [3pt]
    Biodeg     & 1055 $\times$ 41  & 74.41 $\pm$ 2.45 & 76.30 $\pm$ 4.52  & \textbf{77.73} $\pm$ \textbf{3.68} & 74.79 $\pm$ 2.29 \\ [3pt]
    Breast     & 194 $\times$ 33   & 70.77 $\pm$ 6.61 & 71.28 $\pm$ 3.77 & 71.28 $\pm$ 4.41 & \textbf{74.36} $\pm$ \textbf{3.24} \\ [3pt]
    Diabetes   & 768 $\times$ 8    & 72.08 $\pm$ 1.69 & 72.08 $\pm$ 1.69 & \textbf{72.21} $\pm$ \textbf{2.92} & 69.61 $\pm$ 4.07 \\ [3pt]
    Digits     & 3823 $\times$ 63  & 94.72 $\pm$ 0.39 & 94.72 $\pm$ 0.39 & \textbf{94.93} $\pm$ \textbf{0.46} & 93.59 $\pm$ 1.99 \\ [3pt]
    Heart      & 270 $\times$ 25   & 70.74 $\pm$ 3.78 & 70.74 $\pm$ 4.29 & \textbf{71.48} $\pm$ \textbf{5.19} & 67.41 $\pm$ 6.48\\ [3pt]
    Ionosphere & 351 $\times$ 34   & 89.86 $\pm$ 1.64 & \textbf{90.7} $\pm$ \textbf{1.13}  & \textbf{90.7} $\pm$ \textbf{1.13}  & 87.32 $\pm$ 5.19 \\ [3pt]
    Libras     & 360 $\times$ 90   & 95.0 $\pm$ 1.88  & \textbf{95.28} $\pm$ \textbf{1.88} & 95.0 $\pm$ 2.26  & 94.44 $\pm$ 1.76 \\ [3pt]
    Parkinsons & 195 $\times$ 22   & 83.59 $\pm$ 3.48 & \textbf{84.62} $\pm$ \textbf{3.63} & 83.59 $\pm$ 4.47 & 82.05 $\pm$ 6.07 \\ [3pt]
    Phishing   & 11055 $\times$ 68 & 90.75 $\pm$ 0.61 & 90.75 $\pm$ 0.61 & \textbf{91.43} $\pm$ \textbf{0.78} & 90.15 $\pm$ 0.88 \\ [3pt]
    Spam       & 4601 $\times$ 57  & 85.41 $\pm$ 1.72 & 85.99 $\pm$ 1.39 & \textbf{86.47} $\pm$ \textbf{1.13} & 83.71 $\pm$ 2.39 \\ [3pt]
    Spectf     & 267 $\times$ 44   & 75.19 $\pm$ 5.69 & \textbf{78.52} $\pm$ \textbf{3.43} & 78.15 $\pm$ 3.59 & 77.41 $\pm$ 3.78 \\ [3pt]
    Sonar      & 208 $\times$ 60   & 69.52 $\pm$ 6.97 & 70.0 $\pm$ 5.75  & \textbf{72.38} $\pm$ \textbf{6.14} & 66.67 $\pm$ 2.13 \\ [3pt]
    \hline
    \end{tabular}}
    \vspace{1em}

    \caption{Accuracy and Standard Deviation for maximum depth $d = 2$, Gurobi time = 600s}
    \label{table:res2}
\end{table}
\begin{table}[ht]
    
    \centering
    \resizebox{\textwidth}{!}{
    \begin{tabular}{lccccc}   
    \textbf{Dataset} & \textbf{Dimension} & \textbf{CART} & \textbf{TAO} & \textbf{TMO} & \textbf{OCT} \\ [0.5pt]
    \hline \\ 
    A2A & $2265 \times 119$ & 80.79 $\pm$ 1.3  & 80.75 $\pm$ 1.32 & \textbf{80.97} $\pm$ \textbf{1.54} & 78.90 $\pm$ 3.60 \\[3pt]
    A4A & $4781 \times 122$ & 81.9 $\pm$ 1.18  & 82.22 $\pm$ 0.6  & \textbf{82.68} $\pm$ \textbf{0.95} & 80.08 $\pm$ 3.44 \\ [3pt]
    Biodeg & $1055 \times 41$ & 78.1 $\pm$ 1.06  & 78.1 $\pm$ 2.77  & \textbf{79.91} $\pm$ \textbf{1.81} & 74.41 $\pm$ 5.59 \\ [3pt]
    Breast & $ 194 \times 33 $ & 69.74 $\pm$ 5.94 & \textbf{74.36} $\pm$ \textbf{3.63} & 67.69 $\pm$ 3.48 & 72.31 $\pm$ 5.71 \\ [3pt]
    Diabetes & $768 \times 8$ & 69.87 $\pm$ 2.45 & 71.43 $\pm$ 2.63 & \textbf{74.42} $\pm$ \textbf{1.2} & 69.35 $\pm$ 3.97 \\ [3pt]
    Digits & $ 3823 \times 63 $ & 96.44 $\pm$ 0.71 & 96.81 $\pm$ 0.87 & \textbf{96.89} $\pm$ \textbf{0.62} & 92.55 $\pm$ 2.92 \\ [3pt]
    Heart & $ 270 \times 25 $  & 78.15 $\pm$ 0.74 & 78.52 $\pm$ 0.91 & \textbf{81.85} $\pm$ \textbf{2.16} & 79.45 $\pm$ 1.23 \\ [3pt]
    Ionosphere & $ 351 \times 34 $ & \textbf{90.99} $\pm$ \textbf{1.13} & \textbf{90.99} $\pm$ \textbf{0.69} & 87.32 $\pm$ 1.99 & 88.45 $\pm$ 4.22 \\ [3pt]
    Libras & $ 360 \times 90 $ & 93.89 $\pm$ 1.11 & 94.72 $\pm$ 1.36 & \textbf{95.28} $\pm$ \textbf{2.72} & 94.44 $\pm$ 1.76 \\ [3pt]
    Parkinsons & $ 195 \times 22 $ & 81.54 $\pm$ 1.92 & 86.15 $\pm$ 2.61 & \textbf{90.77} $\pm$ \textbf{3.08} & 86.67 $\pm$ 6.36 \\ [3pt]
    Phishing & $ 11055 \times 68 $ & 90.59 $\pm$ 0.56 & 91.23 $\pm$ 0.83 & \textbf{91.92} $\pm$ \textbf{0.45} & 89.43 $\pm$ 0.81 \\ [3pt]
    Spam & $ 4601 \times 57 $ & 87.84 $\pm$ 1.75 & 88.47 $\pm$ 1.5  & \textbf{89.73} $\pm$ \textbf{0.9} & 88.56 $\pm$ 1.23 \\ [3pt]
    Spectf & $ 267 \times 44 $ & 72.59 $\pm$ 9.18 & 72.22 $\pm$ 8.61 & 71.85 $\pm$ 8.23 & \textbf{78.15} $\pm$ \textbf{4.6} \\ [3pt]
    Sonar & $ 208 \times 60 $ & 72.86 $\pm$ 4.9  & \textbf{75.71} $\pm$ \textbf{5.91} & 69.52 $\pm$ 3.81 & 70.95 $\pm$ 3.50 \\ [3pt]
    \hline
    \end{tabular}}
    \vspace{1em}
    
    \caption{Accuracy and Standard Deviation for maximum depth $d = 3$, Gurobi time = 600s}
    \label{table:res3}
\end{table}
\begin{table}[ht]
    \centering

    \resizebox{\textwidth}{!}{
    \begin{tabular}{lccccc}  
    \textbf{Dataset} & \textbf{Dimension} & \textbf{CART} & \textbf{TAO} & \textbf{TMO} & \textbf{OCT} \\ [0.5pt]
    \hline \\ [1pt]
    A2a        & 2265 $\times$ 119 & 80.66 $\pm$ 1.22 & 81.1 $\pm$ 1.44 & \textbf{81.20} $\pm$ \textbf{1.85} & 80.57 $\pm$ 2.84\\ [3pt]
    A4a        & 4781 $\times$ 122 & 82.22 $\pm$ 1.41 & 82.11 $\pm$ 1.27 & \textbf{82.13} $\pm$ \textbf{0.87} & 80.44 $\pm$ 2.89 \\ [3pt]
    Biodeg     & 1055 $\times$ 41  & \textbf{81.23} $\pm$ \textbf{2.26} & 80.85 $\pm$ 2.57 & 80.85 $\pm$ 2.55 & 80.09 $\pm$ 1.44 \\ [3pt]
    Breast     & 194 $\times$ 33   & 67.18 $\pm$ 4.41 & 68.72 $\pm$ 5.23 & 72.31 $\pm$ 5.23 & \textbf{77.44} $\pm$ \textbf{1.03} \\ [3pt]
    Diabetes   & 768 $\times$ 8    & 70.78 $\pm$ 2.36 & 70.65 $\pm$ 3.49 & \textbf{72.21} $\pm$ \textbf{2.07} & 69.61 $\pm$ 4.01 \\ [3pt]
    Digits     & 3823 $\times$ 63  & 97.15 $\pm$ 0.58 & 97.23 $\pm$ 0.48 & \textbf{97.39} $\pm$ \textbf{0.45} & 97.31 $\pm$ 0.28 \\ [3pt]
    Heart      & 270 $\times$ 25   & 75.56 $\pm$ 4.44 & 75.93 $\pm$ 4.22 & \textbf{78.89} $\pm$ \textbf{3.63} & 73.70 $\pm$ 3.78 \\ [3pt]
    Ionosphere & 351 $\times$ 34   & 88.45 $\pm$ 3.26 & \textbf{89.01} $\pm$ \textbf{2.25} & 86.76 $\pm$ 2.61 & 84.51 $\pm$ 3.88 \\ [3pt]
    Libras     & 360 $\times$ 90   & \textbf{95.0} $\pm$ \textbf{1.88} & \textbf{95.0} $\pm$ \textbf{1.11} & 93.61 $\pm$ 3.36 & 94.72 $\pm$ 2.04\\ [3pt]
    Parkinsons & 195 $\times$ 22   & 83.59 $\pm$ 3.48 & 84.1 $\pm$ 2.99 & 85.64 $\pm$ 4.76 & \textbf{86.67} $\pm$ \textbf{5.94} \\ [3pt]
    Phishing   & 11055 $\times$ 68 & 91.67 $\pm$ 0.32 & 91.67 $\pm$ 0.32 & \textbf{92.15} $\pm$ \textbf{0.76} & - \\ [3pt]
    Spam       & 4601 $\times$ 57  & 89.75 $\pm$ 1.17 & 90.16 $\pm$ 1.24 & \textbf{90.47} $\pm$ \textbf{0.68} & 89.88 $\pm$ 1.57 \\ [3pt]
    Spectf     & 267 $\times$ 44   & 72.59 $\pm$ 6.87 & 73.33 $\pm$ 7.46 & 73.33 $\pm$ 6.04 & \textbf{78.52} $\pm$ \textbf{5.44} \\ [3pt]
    Sonar      & 208 $\times$ 60   & \textbf{70.0} $\pm$ \textbf{4.9} & 67.62 $\pm$ 3.56 & 69.52 $\pm$ 3.81 & 68.10 $\pm$ 8.19 \\ [3pt]
    \hline
    \end{tabular}}
    \vspace{1em}
    
    \caption{Accuracy and Standard Deviation for maximum depth $d = 4$, Gurobi time = 600s}
    \label{table:res4}
\end{table}
To assess the performance of our method, we use a total of 14 benchmark datasets from LIBSVM \footnote{\url{https://www.csie.ntu.edu.tw/~cjlin/libsvmtools/datasets/}}  repositories .

The aim of the experiments is to compare our method with respect to three different baselines. The first one is the well known greedy algorithm for decision trees, CART \cite{breiman84classification} which is, even today, the most common approach for classification trees induction. The second one is the alternating optimization TAO algorithm which takes as input the given CART structure and performs a local minimization with respect to the misclassification loss. Finally, we compare our method to the state-of-art MILP approach, OCT, proposed in \cite{bertsimas2017optimal}. 

Each dataset is split up in three parts (64 \% training set, 16 \% validation set, 20 \% test set). We perform experiments considering trees with depth till 4 following the work in \cite{bertsimas2017optimal} that, given the exponential increase of the number of binary variables, it limits the depth to this value.

First, we fit the CART baseline with the standard implementation of the scikit-learn framework \cite{scikit-learn}, setting only the maximum depth parameter and leaving the rest as default. Then, we run a TAO implementation (python3) on the model obtained by scikit-learn to get the second baseline. Finally, we code the OCT formulation of the problem and we use gurobipy \cite{gurobi} to handle the optimization. The original formulation of OCT for axis-aligned splits includes a regularization term to take into account the total number of branch nodes, thus encouraging sparsity in the structure. For this reason, we tune the relative hyperparameter $\alpha$, sampling 3 values from a log-uniform distribution in $[10^{-3}, 1]$ and choosing the one with the best accuracy on the validation set. Moreover, to provide a strong initial upper bound on the optimal solution we follow the idea of the original paper by performing a warm start procedure, initializing the optimization phase with the model obtained by the CART algorithm.

Our method, as described in the previous sections, employs a random forest to initialize the population for the evolutionary procedure. Also in this case we use the standard implementation in scikit-learn to fit the ensemble, setting only the maximum depth and a number of 100 classification trees as the size of the population.

We decide to not validate the cross-rate hyperparameter as we empirically observe that setting $CR = 0.75$ is a good trade-off for the exploration and it allows to obtain competitive solutions. Moreover, with reference to the pseudo-code reported in algorithm \ref{alg:tmo} we set the number of generations of our algorithm $n_{gen}$ equal to 5. Finally, for each method we set the maximum running time limit to 600s.

All the experiments are carried out on a server with an Intel \textregistered  Xeon \textregistered  Gold 6330N CPU with 28 cores and 56 threads @ 2.20GHz. The total available memory is 128GB.
For each dataset we report the size, the mean out-of-sample accuracy and the standard deviation of each method on 5 different seeds.

\section{Discussion}
In tables \ref{table:res2}, \ref{table:res3}, \ref{table:res4} we compare the performance of our approach in case of trees with maximum depth up to 4.
The comparison aims to prove the effectiveness of our method with respect to different depths, to capture also information about the performance trend with respect to the transparency of the model.
Moreover, the main contribution is related to the applicability of our method to real world datasets with thousand of samples in opposition to the MILP approach that often fails to obtain a better solution than the one provided by the CART warm start.

Both for depth 2 and 3, results highlights the effectiveness of our memetic evolutionary method which outperforms other approaches on 9 out of 14 datasets (64\%). More generally, it is evident that TMO is often able to induce structures with better generalization capabilities even for datasets with thousands of samples. This result is particularly evident for larger datasets (Phishing, A4A, Spam, Digits) where MILP models suffers the fact that the number of binary variables scales exponentially with respect to the number of samples.
Moreover, OCT, although it exploits the warm start procedure, can rarely outperform CART. This drawback is mainly related to the model becoming intractable, making hard for the MILP solver to find any better feasible solution than the one initially provided by CART.

An example of this fact is visible in the results for depth 4 (table \ref{table:res4}) for the phishing dataset. The OCT approach exceeds the availbale RAM (128GB) as the branch and bound tree allocated by the Guroby solver requires too much memory, causing the operating system to kill the process. This is clearly a consequence of the $\mathcal{NP}$-completeness of the problem that makes exact models unusable for real world datasets even with small trees.

These latest results show also that OCT can sometimes provide structures with better out-of-sample accuracy for deeper trees, exploiting the regularization term that may discover pruned models, encouraging generalization.
However, also at this depth, our algorithm is the most competitive among the four proposed, achieving the best performance on half of the datasets.

Finally, although theoretically able to provide the certified global optimum, in practice results show that this optimal solution is almost never reached (or at least certified) because of the difficulties in closing the optimality gap. For this reason, our method, with no claim to provide the global optimum, has proven to be a good trade-off between the applicability of over-exploited greedy approaches and the need of exploration of the feasible space to induce near-optimal solutions.

\section{Computational complexity}
Unlike MILP models, the complexity of our method does not depend exponentially by the number of samples. In fact, at each iteration of the algorithm, the method performs the selection/crossover phase which can be made in $O(2^d)$, then an instance of the local optimizer TAO is performed and its cost is comparable to running CART to grow a tree of the same depth \cite{carreira2018alternating}, thus getting a complexity of $O(npd)$.
These two routines are repeated for each tree in the population getting a total cost of $O(k2^d + knpd)$.
Note that in case of shallow trees and large datasets, complexity mainly depends on the term $O(knpd)$ saying that a generation of TMO is comparable to running CART a number of times equal to the size of the population.


\section{Conclusion}\label{sec5}
We proposed TMO, a novel memetic algorithm for classification trees induction which is able to discover structures with generalization capabilities that are competitive to the state of art algorithms for CTs growing.

We employ the standard selection/crossover procedure of evolutionary algorithms to explore the space of feasible classification trees for a given depth. Moreover, the use of the memetic strategy was found to be decisive in defining good quality models. Numerical experiments suggest that the main contribution of our method is the applicability to real world datasets with thousands of samples, reaching a competitive accuracy even when exact approaches fail to find better solutions than the one provided in warm start.

A critical difference between other algorithms is the way TMO explores the feasible space. Our method, in fact, searches for structures using directly the misclassification loss, thus not reintroducing impurity measures that were used to define the initial population.

Our work opens up to several extensions such as regression trees and multivariate trees. Moreover, new techniques can be defined to maintain in the population good sub-trees that may acts as upper bound solutions to speed up the search, pruning the feasible space.

\let\textbf\relax
\bibliography{bibliography.bib}




\end{document}